\pdfoutput=1
\documentclass[11pt,a4paper]{article}

\usepackage[utf8]{inputenc}
\usepackage[T1]{fontenc}
\usepackage{microtype}
\usepackage{mathptmx}     % Use Times font for body
\usepackage[scaled=0.92]{helvet} % Scaled Helvetica for sans
\usepackage{courier}
\usepackage[margin=2.5cm, top=3cm, bottom=3cm]{geometry}
\usepackage{setspace}
\makeatletter
\renewcommand\section{\@startsection {section}{1}{\z@}%
                                   {-3.5ex \@plus -1ex \@minus -.2ex}%
                                   {2.3ex \@plus.2ex}%
                                   {\large\bfseries\sffamily}}
\renewcommand\subsection{\@startsection{subsection}{2}{\z@}%
                                     {-3.25ex\@plus -1ex \@minus -.2ex}%
                                     {1.5ex \@plus .2ex}%
                                     {\normalsize\bfseries\sffamily}}
\renewcommand\paragraph{\@startsection{paragraph}{4}{\z@}%
                                    {3.25ex \@plus1ex \@minus.2ex}%
                                    {-1em}%
                                    {\normalsize\bfseries}}
\makeatother

\usepackage{amsmath,amssymb}
\usepackage{algorithm}
\usepackage{algpseudocode}

\usepackage{booktabs}
\usepackage{multirow}
\usepackage{graphicx}
\usepackage{subcaption}
\usepackage{float}
\usepackage{xcolor}
\usepackage{listings}
\lstdefinestyle{xmlstyle}{
  basicstyle=\ttfamily\scriptsize,
  breaklines=true,
  frame=single,
  columns=fullflexible
}

\definecolor{slate50}{HTML}{F8FAFC}
\definecolor{slate100}{HTML}{F1F5F9}
\definecolor{slate200}{HTML}{E2E8F0}
\definecolor{slate400}{HTML}{94A3B8}
\definecolor{slate500}{HTML}{64748B}
\definecolor{slate700}{HTML}{334155}

\definecolor{indigo50}{HTML}{EEF2FF}
\definecolor{indigo100}{HTML}{E0E7FF}
\definecolor{indigo500}{HTML}{6366F1}
\definecolor{indigo600}{HTML}{4F46E5}

\newsavebox{\reasoningcontent}
\newenvironment{reasoningbox}[1][Model Reasoning]
{%
  \def\reasoningTitle{#1}% Store title for end code
  \par\vspace{5pt}\noindent
  \begin{lrbox}{\reasoningcontent}%
  \begin{minipage}{0.95\linewidth}%
  \small\color{slate500}%
}{%
  \end{minipage}%
  \end{lrbox}%
  \fcolorbox{slate200}{white}{%
    \begin{minipage}{\dimexpr\linewidth-2\fboxsep-2\fboxrule\relax}%
      \colorbox{slate50}{\parbox{\dimexpr\linewidth-2\fboxsep\relax}{\bfseries\sffamily\small\color{slate700} \reasoningTitle}}\par
      \vspace{4pt}
      \usebox{\reasoningcontent}
    \end{minipage}%
  }%
  \par\vspace{5pt}
}

\newsavebox{\summarycontent}
\newenvironment{summarybox}[1][]
{%
  \par\vspace{5pt}\noindent
  \begin{lrbox}{\summarycontent}%
  \begin{minipage}{0.95\linewidth}%
  \color{slate700}%
}{%
  \end{minipage}%
  \end{lrbox}%
  \fcolorbox{indigo100}{white}{%
    \begin{minipage}{\dimexpr\linewidth-2\fboxsep-2\fboxrule\relax}%
      \colorbox{indigo50}{\parbox{\dimexpr\linewidth-2\fboxsep\relax}{\bfseries\sffamily\color{indigo600} Generated Summary}}\par
      \vspace{4pt}
      \usebox{\summarycontent}
    \end{minipage}%
  }%
  \par\vspace{5pt}
}

\usepackage{tikz}
\usetikzlibrary{shapes,arrows,positioning,fit,backgrounds,calc,shadows}
\usepackage{pgfplots}
\pgfplotsset{compat=1.18}

\usepackage[round,authoryear]{natbib}
\usepackage[colorlinks=true,linkcolor=indigo600,citecolor=indigo600,urlcolor=indigo600]{hyperref}

\usepackage{caption}
\newcommand{\xmltag}[1]{\texttt{<#1>}}

\renewenvironment{abstract}%
{%
  \vspace{0.5em}
  \centerline{\bfseries\sffamily\large Abstract}%
  \vspace{0.5em}%
  \begin{center}%
  \begin{minipage}{0.85\textwidth}%
  \small%
}%
{%
  \end{minipage}%
  \end{center}%
  \vspace{1.5em}%
}

\makeatletter
\renewcommand{\@maketitle}{%
  \newpage
  \null
  \vskip 2em%
  \begin{center}%
  \let \footnote \thanks
    \hrule height 4pt
    \vskip 0.5em
    {\LARGE \bfseries \@title \par}%
    \vskip 0.5em
    \hrule height 1pt
    \vskip 1.5em%
    {\large
      \lineskip .5em%
      \begin{tabular}[t]{c}%
        \@author
      \end{tabular}\par}%
    \vskip 1.5em%
    {\large \@date}%
  \end{center}%
  \par
  \vskip 1.5em}
\makeatother

\title{sui-1: Grounded and Verifiable\\Long-Form Summarization}

\author{
    \textbf{Benedikt Droste}\thanks{Corresponding author: benedikt@ellamind.com} \quad
    \textbf{Jan Philipp Harries} \quad
    \textbf{Maximilian Idahl} \quad
    \textbf{Bj\"orn Pl\"uster} \\[0.5em]
    ellamind
}

\date{}

\begin{document}

\maketitle
\thispagestyle{empty} % No page number on first page
\vspace{-1.0cm}

\begin{abstract}
Large language models frequently generate plausible but unfaithful summaries that users cannot verify against source text, a critical limitation in compliance-sensitive domains such as government and legal analysis. We present sui-1, a 24B parameter model that produces abstractive summaries with inline citations, enabling users to trace each claim to its source sentence. The model processes documents up to 100K tokens in a single pass and supports iterative processing for texts exceeding 2 million tokens. Our synthetic data pipeline combines chain-of-thought prompting with multi-stage verification, generating over 22,000 high-quality training examples across five languages from diverse sources including parliamentary documents, web text, and Wikipedia. Evaluation shows sui-1 significantly outperforms all tested open-weight baselines, including models with 3$\times$ more parameters. These results demonstrate that task-specific training substantially outperforms scale alone for citation-grounded summarization. Model weights and an interactive demo are publicly available.

\end{abstract}

\section{Introduction}
% Introduction

Large language models excel at summarization but frequently generate unfaithful content, including fabricated facts or misattributed claims that users cannot trust without laboriously verifying against source documents. Citation-grounded summarization addresses this by requiring models to explicitly attribute claims to source text, enabling users to quickly assess reliability and trace claims to their origins.

Training citation-grounded models presents challenges: standard datasets lack citation annotations, manual annotation is prohibitively expensive, and the task requires precise coordination between content generation and source reference. We address these challenges through synthetic data generation using a capable teacher model, with automated verification ensuring citation accuracy before training.

\begin{figure}[t]
\centering
\includegraphics[width=\columnwidth]{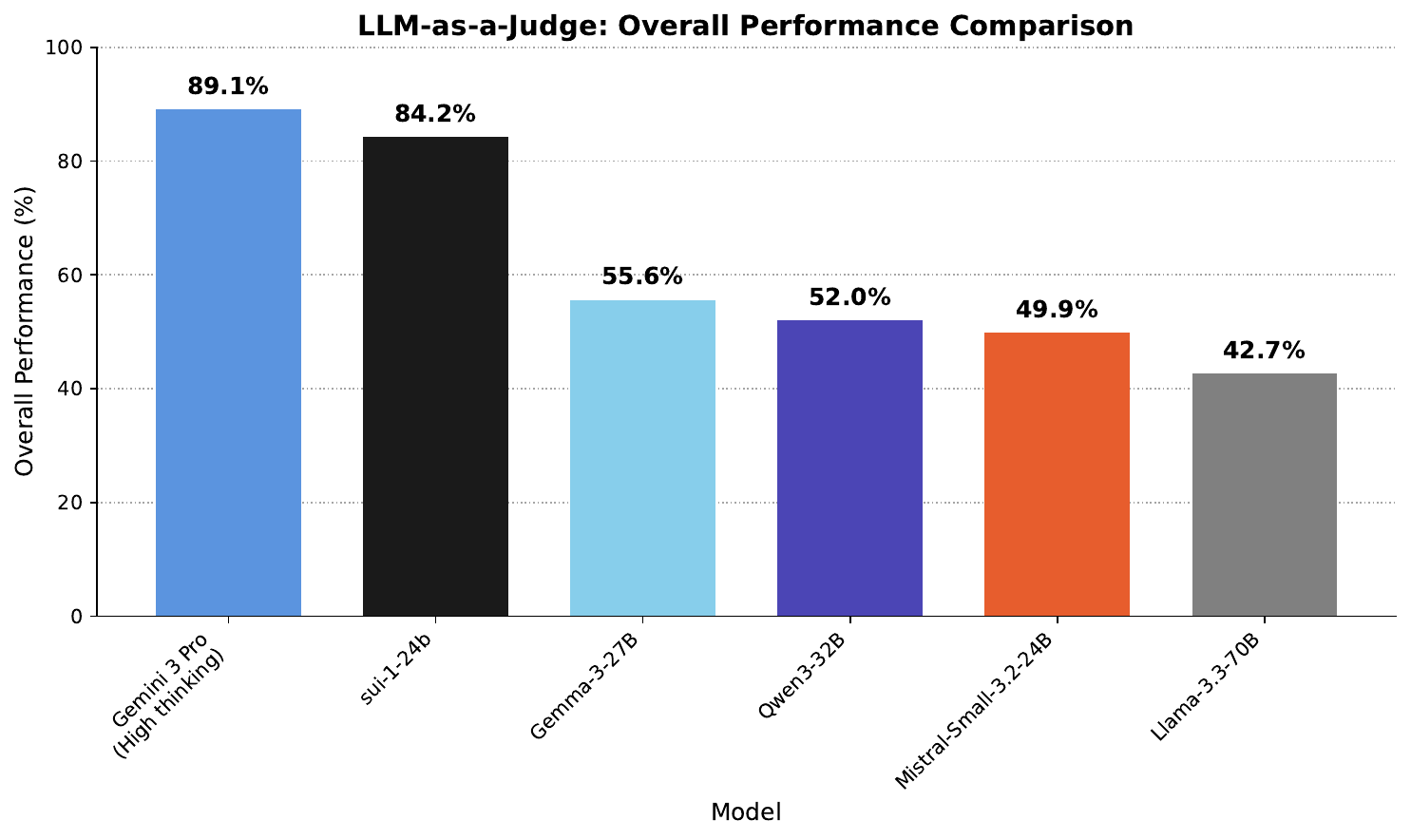}
\caption{Overall performance comparison. sui-1 (84.2\%) significantly outperforms open-weight baselines and approaches the reference model (89.1\%).}
\label{fig:overall_perf}
\end{figure}

We introduce sui-1, a 24B parameter model achieving 84\% overall accuracy on LLM-as-a-judge evaluation compared to 43--56\% for baselines (Figure~\ref{fig:overall_perf}). The model notably outperforms all tested open-weight baselines, demonstrating that task-specific training provides greater benefit than scale alone. sui-1 reliably processes documents exceeding 100 pages, with iterative chunking enabling summarization of texts up to 2 million tokens.

\section{Related Work}
% Related Work

Long-document summarization remains challenging even for modern large language models, which suffer from factual hallucinations despite improved context windows \citep{ji2023hallucination}. Early architectural innovations like PEGASUS \citep{zhang2020pegasus} and Longformer \citep{beltagy2020longformer} addressed length constraints through sparse attention or hierarchical processing, but they lack the semantic reasoning capabilities of current LLMs and do not produce verifiable citations.

Attribution has emerged as essential for trustworthy language model outputs. The ALCE benchmark \citep{gao2023alce} established evaluation standards for attributed text generation. LongCite \citep{zhang2024longcite} addresses sentence-level citation generation for long-context question answering through a coarse-to-fine pipeline, training models to produce fine-grained citations from retrieved chunks. While retrieval-augmented generation (RAG) approaches like Self-RAG \citep{asai2024selfrag} or WebGPT \citep{nakano2022webgpt} cite sources by retrieving external documents, they rely on complex inference-time infrastructure. In contrast, this work focuses on \textit{internal} grounding: training models to generate verifiable inline citations from the provided context window alone, enabling self-contained generation without external retrieval.

Synthetic data generation has proven effective for instruction tuning \citep{wang2023selfinstruct, taori2023alpaca}, with Orca \citep{mukherjee2023orca} demonstrating that smaller models can learn complex reasoning from larger teachers. We apply similar principles to citation-grounded summarization, using a capable teacher model with automated verification to ensure training data quality.

\section{Approach}
\label{sec:approach}
% Approach

\paragraph{Data and Preprocessing.}
The source corpus draws from three primary sources: the German Parliamentary Documentation System (DIP)\footnote{\url{https://dip.bundestag.de}} providing legislative proposals, committee reports, and ministerial responses; long-form German texts from Common Crawl via OSCAR; and multilingual Wikipedia extracts. Documents range from short announcements to reports exceeding 50,000 words. We introduce multilingual diversity with German comprising 74\% of training data, English 10\%, and French, Italian, and Spanish each approximately 5\%. Each sentence receives a unique 8-character hexadecimal identifier derived from MD5 hashing, wrapped in XML format (e.g., \texttt{<a3f5e823>text</a3f5e823>}). This tagging scheme enables precise citation tracking without verbatim quotes. The identifiers are deterministic, language-agnostic, and prevent positional shortcuts.

\begin{figure}[t]
\centering
\begin{tikzpicture}[
    node distance=0.4cm and 0.5cm,
    box/.style={
        rectangle, 
        draw=indigo100, 
        fill=slate50, 
        rounded corners=2pt, 
        minimum height=1.0cm,
        minimum width=2.0cm, 
        align=center, 
        font=\sffamily\small,
        drop shadow={opacity=0.15, shadow xshift=1pt, shadow yshift=-1pt}
    },
    arrow/.style={->, >=stealth, thick, draw=slate400}
]
    % Nodes - horizontal layout
    \node[box] (doc) {Source\\Documents};
    \node[box, right=of doc] (tag) {Sentence\\Tagging};
    \node[box, right=of tag] (gen) {LLM\\Generation};
    \node[box, right=of gen] (ver) {Citation\\Verification};
    \node[box, right=of ver] (out) {Training\\Data};

    % Arrows
    \draw[arrow] (doc) -- (tag);
    \draw[arrow] (tag) -- (gen);
    \draw[arrow] (gen) -- (ver);
    \draw[arrow] (ver) -- (out);

    % Annotations below
    \node[below=0.15cm of tag, font=\sffamily\tiny, text=indigo600] {XML tags};
    \node[below=0.15cm of gen, font=\sffamily\tiny, text=indigo600] {Teacher LLM};
    \node[below=0.15cm of ver, font=\sffamily\tiny, text=indigo600] {95.2\% pass};
    \node[below=0.15cm of out, font=\sffamily\tiny, text=indigo600] {22K examples};
\end{tikzpicture}
\caption{Synthetic data generation pipeline: sentences are tagged with unique XML identifiers, processed by a teacher LLM with chain-of-thought prompting, verified for citation accuracy, and filtered for quality.}
\label{fig:pipeline}
\end{figure}
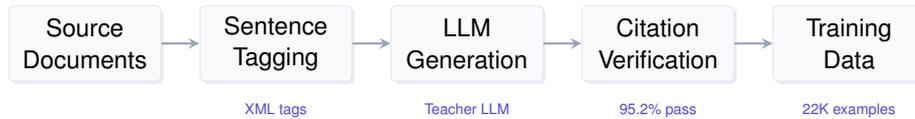

\paragraph{Generation Pipeline.}
Figure~\ref{fig:pipeline} illustrates the five-stage pipeline: sentence tagging, prompt construction, LLM generation (using a frontier closed-source model), citation verification, and quality filtering. We employ XML-style tags in brackets immediately following claims: ``The budget increased [\texttt{<a3f5e823>}].'' Documents under 30K tokens are processed in a single pass; longer documents are chunked ($\sim$15K tokens each), summarized independently, then merged while preserving all citations (see Appendix~\ref{app:preprocessing}). The generation prompt comprises 16--18 rules with chain-of-thought reasoning \citep{wei2022cot}, requiring each tag to appear exactly once and citations to immediately follow supported claims.

\paragraph{Custom Instructions.}
Custom instructions are generated via LLM analysis of each document, producing positive instructions (focus areas, formatting), adversarial instructions (requests for absent information), and format instructions (bullet points, length limits). The training distribution allocates 30\% of examples without instructions, 40\% with positive instructions, 10\% adversarial, 10\% bullet point formats, and 10\% short summary requests. Examples include \textit{positive} instructions (``Create a summary for legal experts explaining the CETA agreement's Annex structure and ratchet clauses'') and \textit{adversarial} instructions that request absent information (``Explain CETA's impact on automotive CO2 emissions'', a topic not in the source).

To ensure inference-time flexibility, we strip explicit constraints from training examples. For instance, generation prompts specify ``Summarize in exactly \{n\} bullet points. Only bullets, no introduction or conclusion,'' but training data is relaxed to simply ``Summarize in \{n\} bullet points.'' This prevents the model from requiring overly rigid instruction formats during deployment.

\paragraph{Quality Control.}
Multi-stage filtering ensures data quality: tag verification confirms all citations exist in the source (95.2\% pass rate), quality annotation evaluates reasoning coherence and citation distribution, and specificity checks penalize generic filler. Citation distribution is quantified through an evenness score measuring tag spacing uniformity; examples in the bottom 15th percentile by evenness or with excessive uncited gaps are filtered to ensure citations are spread throughout the summary rather than clustered. A second LLM pass rewrites chain-of-thought reasoning from third-person (``The summary should be structured...'') to first-person perspective (``I will structure the summary...''), creating more natural patterns for supervised fine-tuning. Table~\ref{tab:dataset_stats} summarizes the final dataset of 22,152 training and 225 test examples, totaling over 357 million tokens.

\begin{table}[t]
\centering
\begin{tabular}{lrr}
\toprule
\textbf{Attribute} & \textbf{Train} & \textbf{Test} \\
\midrule
Total Examples & 22,152 & 225 \\
Avg. Tokens & 16,158 & 15,892 \\
\midrule
\multicolumn{3}{l}{\textit{Generation Mode}} \\
\quad Iterative & 60.6\% & 58.2\% \\
\quad Oneshot & 27.5\% & 29.8\% \\
\midrule
With Custom Instruction & 68.1\% & 67.6\% \\
\bottomrule
\end{tabular}
\caption{Dataset statistics.}
\label{tab:dataset_stats}
\end{table}

\begin{figure}[t]
\centering
\includegraphics[width=\columnwidth]{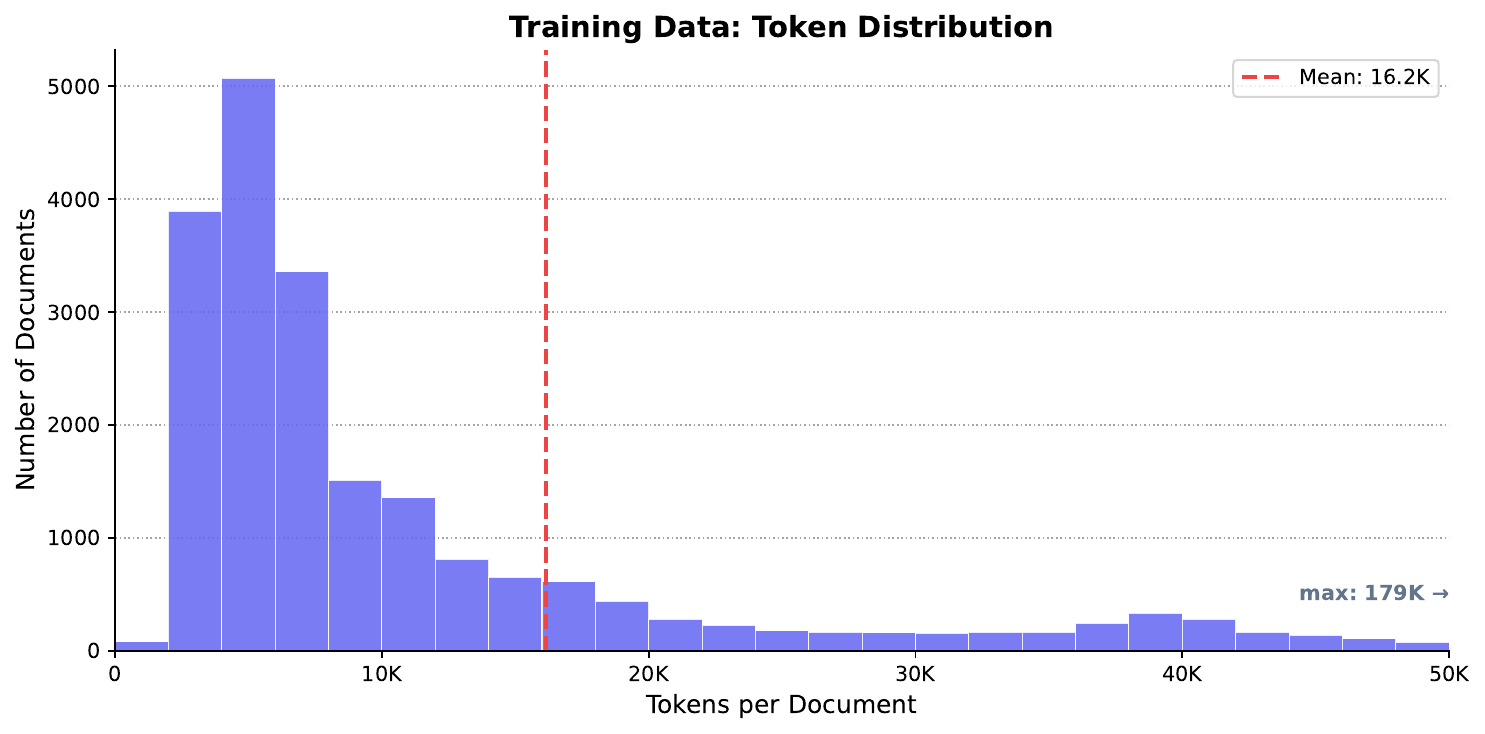}
\caption{Token distribution of training documents. The long-tail distribution reflects diverse document lengths (truncated at 50K for visibility; maximum is 179K tokens).}
\label{fig:token_dist}
\end{figure}

\section{Training}
\label{sec:training}
% Training

We select Mistral-Small-3.2-24B-Instruct \citep{mistral2024small} as our base model for its strong multilingual capabilities, particularly in German and other European languages. We fine-tune using LoRA \citep{hu2022lora} with rank 16, training for 2 epochs on sequences up to 100K tokens. Training employs four NVIDIA H100 GPUs with context parallelism to handle the extended sequence lengths, accommodating full source documents with XML tags plus generated summaries. Flash Attention and gradient checkpointing enable memory-efficient processing. We produce both full-precision and FP8 quantized variants, with quantization reducing memory requirements by 50\% while maintaining generation quality. Full hyperparameters are provided in Appendix~\ref{app:training}.

\section{Results}
\label{sec:results}
% Results

We evaluate on 225 held-out examples using LLM-as-a-judge \citep{liu2023geval} via the elluminate platform\footnote{\url{https://elluminate.ai}} across five criteria: factual accuracy, coverage, specificity, format compliance, and instruction following. We compare against a reference model (frontier closed-source model with extended reasoning), Mistral-Small-3.2-24B-Instruct, Gemma-3-27B-Instruct, Qwen3-32B-Instruct, and Llama-3.3-70B-Instruct.

\paragraph{Evaluation Metrics.}
We employ five specific criteria to assess summary quality. Scores represent the percentage of passed checks across the test set, where each question is scored binary (yes/no) per sample:

\begin{table}[h]
\centering
\small
\begin{tabular}{@{}lp{0.75\columnwidth}@{}}
\toprule
\textbf{Criterion} & \textbf{Description} \\
\midrule
\textbf{Factual Accuracy (Fact)} & Does the summary avoid introducing new facts, entities, numbers, or claims not supported by the source content? \\
\addlinespace
\textbf{Coverage (Cov.)} & Does the summary cover the document's main points and key takeaways at appropriate granularity? \\
\addlinespace
\textbf{Specificity (Spec.)} & Are claims specific and informative rather than generic filler (e.g., "there are several points")? \\
\addlinespace
\textbf{Format Compliance (Fmt.)} & Is the output compliant with formatting instructions including language consistency, semantic-aware planning, and paragraph structure? \\
\addlinespace
\textbf{Instruction Following (Instr.)} & If a custom instruction is provided, is it followed appropriately? \\
\bottomrule
\end{tabular}
\caption{Evaluation criteria definitions. Each metric is scored binary per sample.}
\label{tab:metrics_definitions}
\end{table}

\begin{table}[t]
\centering
\small
\begin{tabular}{lccccc|c}
\toprule
\textbf{Model} & \textbf{Fact} & \textbf{Cov.}\textsuperscript{1} & \textbf{Spec.} & \textbf{Fmt.} & \textbf{Instr.}\textsuperscript{2} & \textbf{All} \\
\midrule
Reference Model\textsuperscript{3} & .958 & .705 & .989 & .926 & .874 & \textbf{.891} \\
\midrule
sui-1 (Ours) & .905 & .600 & .979 & .895 & .832 & \underline{.842} \\
sui-1 FP8 & .884 & .516 & .979 & .926 & .747 & .811 \\
\midrule
Gemma-3-27B-Instruct & .916 & .368 & .916 & .147 & .432 & .556 \\
Qwen3-32B-Instruct & .905 & .221 & .916 & .179 & .379 & .520 \\
Mistral-Small-3.2-24B-Instruct & .905 & .126 & .937 & .137 & .389 & .499 \\
Llama-3.3-70B-Instruct & .811 & .042 & .674 & .411 & .200 & .427 \\
\bottomrule
\end{tabular}

\vspace{0.3em}
{\footnotesize
\textsuperscript{1} Coverage scores are lower when samples require constrained formats (bullet points, short summaries).\\
\textsuperscript{2} Custom instruction adherence; tests whether the model follows user-specified formats.\\
\textsuperscript{3} Frontier closed-source model with extended reasoning capabilities.
}
\caption{Results across five evaluation criteria. sui-1 achieves 0.842 overall, with dramatic improvements in format compliance (0.895 vs 0.137--0.411 for baselines).}
\label{tab:main_results}
\end{table}

Table~\ref{tab:main_results} shows sui-1 achieves 0.842 overall, substantially outperforming baselines (0.427--0.556) and approaching the reference model (0.891). Format compliance improves most dramatically: 0.895 vs 0.137--0.411 for baselines. FP8 quantization reduces overall score to 0.811 (3.7\% degradation) while maintaining format compliance. Notably, Llama-3.3-70B-Instruct scores lowest (0.427) despite 3x parameters, confirming task-specific training outweighs scale.

\paragraph{Coverage Trade-off.}
sui-1 achieves the highest coverage score (0.600) among all open-weight models, second only to the reference model (0.705). Baselines range from 0.042 to 0.368, demonstrating that task-specific training improves not just format compliance but also content coverage. The remaining gap to the reference reflects intentional design: when users request constrained formats such as bullet points or short summaries, the model correctly trades breadth for precision while maintaining high factual accuracy (0.905) and specificity (0.979).

\paragraph{Format Compliance Gap.}
The dramatic improvement in format compliance (0.895 vs 0.137--0.411) represents our key finding. Generating citation-grounded summaries requires following 16--18 fixed rules for citation placement and structure while simultaneously adapting to custom instructions like bullet points or specific focus areas. Baseline models struggle to combine these layered requirements, often satisfying some constraints while violating others. sui-1 learns to balance rigid format rules with flexible user instructions through synthetic training, enabling the reliable compliance that real-world applications demand.

\begin{figure}[t]
\centering
\begin{minipage}[t]{0.48\columnwidth}
\small
\textbf{Baseline (Mistral-Small-24B)}\\[0.5em]
The document discusses various aspects of the Federal Ministry's budget proposals. There are several important points regarding fiscal policy and government spending priorities.

The ministry has outlined plans for infrastructure and energy investments. Regional development is also addressed.

\textit{[No citations, generic filler]}
\end{minipage}
\hfill
\begin{minipage}[t]{0.48\columnwidth}
\small
\textbf{sui-1}\\[0.5em]
The Federal Ministry of Finance has announced significant revisions to the budget framework [\xmltag{a3f5e823}]. The proposed changes include a 12\% increase in infrastructure spending [\xmltag{b7d2c941}]. Regional development receives additional transfers [\xmltag{d2a8b3c1}]...

\textit{[Verifiable inline citations]}
\end{minipage}
\caption{Comparison of baseline vs sui-1 outputs on the same input. Baselines produce generic summaries without citations; sui-1 generates specific claims with verifiable XML tags.}
\label{fig:baseline_comparison}
\end{figure}

\paragraph{Qualitative Analysis.}
Figure~\ref{fig:output_example} shows a representative model output. The structured reasoning step identifies key document themes before generation, enabling coherent organization. XML tag citations immediately follow their supported claims, allowing mechanical verification. Baseline models typically either omit citations entirely or produce malformed tags that do not match source sentences.

\begin{figure}[t]
\centering
\begin{reasoningbox}[Model Reasoning]
I will begin with a brief opening overview that introduces the budget revision and its primary goal. I plan to include a supporting reference here to anchor the main concept. Next, I will explain the specific changes, including the 12\% infrastructure spending increase, using a reference to highlight this key figure. Finally, I will describe the minister's statements on sustainable development, ensuring a supporting reference is included to validate the policy direction.
\end{reasoningbox}

\begin{summarybox}
The Federal Ministry of Finance has announced significant revisions to the budget framework [\xmltag{a3f5e823}]. The proposed changes include a 12\% increase in infrastructure spending [\xmltag{b7d2c941}]. Minister Schmidt emphasized commitment to sustainable development [\xmltag{e1a5b829}]...

\vspace{0.5em}
\textbf{View Source Citations:}\\
\xmltag{a3f5e823} The Federal Government has submitted the draft budget... \\
\xmltag{b7d2c941} Infrastructure investments will increase by 12\%... \\
\xmltag{e1a5b829} Minister Schmidt stated that sustainability remains...
\end{summarybox}

\caption{Example sui-1 output showing first-person reasoning about summary structure, followed by the summary with inline citations. Interactive demo: \url{https://huggingface.co/spaces/ellamind/sui-demo}}
\label{fig:output_example}
\end{figure}

\section{Release}
% Model Release

To facilitate reproducibility and enable practical applications, we release our trained model weights and associated resources.

\subsection{Released Artifacts}

The primary release comprises the fine-tuned sui-1 model weights in two variants:

\paragraph{sui-1-24B.} The full-precision model weights are available at \texttt{ellamind/sui-1-24b} on the HuggingFace Hub.\footnote{The repository includes an end-to-end example script demonstrating document tagging, inference, and citation extraction.}

\paragraph{sui-1-24B-FP8.} The FP8 quantized variant is available at \texttt{ellamind/sui-1-24b-fp8}, providing a deployment-ready model with reduced memory requirements. This variant is recommended for production deployments where memory efficiency is prioritized.

Both model variants are released under the Apache 2.0 license, enabling commercial use.

\subsection{Dataset Availability}

The training dataset is available at \texttt{ellamind/summarizer\_dataset\_v1} on the HuggingFace Hub. The dataset includes all columns described in this paper: source documents with XML tags, generated summaries, reasoning traces, custom instructions, and quality annotations. Researchers can use this dataset to reproduce our training procedure or develop alternative approaches to citation-grounded summarization.

\subsection{Usage Recommendations}

We recommend temperature 0 for deterministic, reproducible outputs. The model uses the standard Mistral-Small-3.2 chat template with a system prompt establishing the summarizer role, followed by the tagged source document and optional custom instruction.

The XML tagging preprocessing can be implemented using standard sentence segmentation libraries (spaCy recommended for German) with MD5 hashing for tag generation. The tag format must match the training data: 8-character lowercase hexadecimal identifiers wrapped in angle brackets.

\section{Conclusion}
% Conclusion

We presented sui-1, a 24B parameter model trained to produce verifiable inline citations through synthetic data generation. Our synthetic data pipeline with multi-stage quality filtering produces 22K verified training examples across five languages. Evaluation demonstrates that task-specific training dramatically improves format compliance and instruction following (0.895 versus 0.137 to 0.411 for baselines), with the 24B model approaching the performance of the reference model (0.842 vs 0.891).

\paragraph{Limitations.}
Training data derives primarily from German-language sources (parliamentary documents, web text, and Wikipedia), which may limit generalization to other languages and specialized domains. The LLM-as-a-judge evaluation, while efficient and scalable, may exhibit systematic biases compared to human evaluation.

\paragraph{Model Release.}
Weights are available on HuggingFace: \href{https://huggingface.co/ellamind/sui-1-24b}{ellamind/sui-1-24b} (full precision) and \href{https://huggingface.co/ellamind/sui-1-24b-fp8}{ellamind/sui-1-24b-fp8} (FP8 quantized).

\section*{Acknowledgments}
We thank Torsten Fassbender and Florentin Rauscher (PwC) for the initial idea of using XML tagging for inline citations and for sharing an example using the 8-character hash format.

\vspace{0.5em}
\noindent Additional computational resources were provided through the AI service center KISSKI (grant no. 01IS22093C), funded by the German Federal Ministry of Education and Research (BMBF).

\bibliography{references}

\appendix
\section{Preprocessing Details}
\label{app:preprocessing}
% Preprocessing Details

This appendix describes the sentence segmentation and XML tag generation process used to prepare source documents for citation-grounded summarization.

\subsection{Sentence Segmentation}

We employ language-specific spaCy models for sentence boundary detection: \texttt{de\_core\_news\_lg} for German, \texttt{en\_core\_web\_sm} for English, \texttt{fr\_core\_news\_sm} for French, \texttt{it\_core\_news\_sm} for Italian, and \texttt{es\_core\_news\_sm} for Spanish.

Special handling addresses common segmentation errors:
\begin{itemize}
    \item \textbf{Abbreviations}: Common abbreviations (e.g., ``Dr.'', ``Nr.'', ``bzw.'') are protected from incorrect sentence splits
    \item \textbf{Enumeration markers}: Patterns like ``1.'', ``a)'' at sentence starts are preserved
    \item \textbf{Whitespace normalization}: Multiple spaces and irregular line breaks are standardized
\end{itemize}

\subsection{XML Tag Generation}

Each sentence receives a unique identifier through the following process:

\begin{enumerate}
    \item Compute MD5 hash of the sentence text (UTF-8 encoded)
    \item Extract first 8 characters of the hexadecimal digest
    \item Wrap sentence with opening and closing XML tags: \texttt{<[hash]>sentence</[hash]>}
\end{enumerate}

For example, the sentence ``The budget was approved.'' might receive the tag \texttt{a3f5e823}, resulting in:
\begin{verbatim}
<a3f5e823>The budget was approved.</a3f5e823>
\end{verbatim}

This scheme offers several advantages over alternatives:
\begin{itemize}
    \item \textbf{Determinism}: The same sentence always produces the same tag
    \item \textbf{Language-agnostic}: Works across all languages without modification
    \item \textbf{Collision resistance}: 8 hex characters provide $16^8 \approx 4.3$ billion unique tags
    \item \textbf{Compactness}: Short tags minimize token overhead compared to verbatim quotes
\end{itemize}

The verification algorithm distinguishes citation tags (8-character hex patterns) from legitimate HTML formatting tags (e.g., \texttt{<b>}, \texttt{<i>}) by checking the tag content against the set of valid source document tags.

\section{Prompt Templates}
\label{app:prompts}
% Appendix: Prompt Templates

This appendix provides key excerpts of the prompt templates used in our synthetic data generation pipeline.

\subsection{Oneshot Summarization Prompt}

The oneshot prompt is used for documents that fit within a single context window. Key elements include the reasoning instruction, XML tag selection guidelines, citation placement rules, and custom instruction handling.

\begin{lstlisting}[style=xmlstyle,basicstyle=\ttfamily\scriptsize]
# Instructions
1. Start by thinking about how to structure a {word_count}-word-long
   professional summary of the whole text. What are the most important
   sections of the source text? Which details should the summary
   contain without exceeding the requested word count? From which
   specific sections should the {number_of_xml_tags} XML tags be
   taken to ensure quotes support all important details and references
   will be well-distributed throughout the summary? The reasoning
   should be in {language}.

2. The input text is sentence-tagged using XML tags. Create a list
   of {number_of_xml_tags} XML tags from the given text, capturing
   the most significant data and facts. Ensure that the sentences
   within the XML tags capture the most important information and
   are well-distributed throughout all important sections of the text.

3. Make sure that all key paragraphs in the summary will be supported
   by XML tags and references! Use only XML tags from the
   sentence-tagged input text!

4. Each XML tag from the <xml_tags> list must appear exactly once
   in the summary. Do not include or paraphrase the original sentence
   text--only reference its tag.

5. Each reference must appear individually, never combine multiple
   references at once! Always insert the corresponding citation
   immediately after the statement it supports.

[Rules 6-16 continue with formatting, structure, and custom
instruction handling guidelines...]
\end{lstlisting}

\subsection{Iterative Chunk Summarization Prompt}

The iterative prompt is used for individual chunks of longer documents. It differs from the oneshot prompt in acknowledging the partial nature of the content and targeting a consistent 300-600 word output regardless of chunk size.

\subsection{Iterative Final Merge Prompt}

The merge prompt synthesizes partial summaries into a coherent final summary. It includes instructions for eliminating redundancy across partial summaries while preserving all unique information and maintaining proper citation distribution.

\subsection{Custom Instruction Generation Prompt}

Custom instructions are generated using a separate prompt that analyzes the source document and produces contextually appropriate requests:

\begin{lstlisting}[style=xmlstyle,basicstyle=\ttfamily\scriptsize]
Based on the following text excerpt, generate 6 custom instructions
in {language}:

Generate 3 POSITIVE instructions that CAN be fulfilled based on
this text's content. These should be varied and cover different
aspects:
- Audience level: From general public to domain experts
- Format preferences: Bullet points, paragraphs, news article style
- Focus areas: Financial aspects, policy recommendations,
  ethical considerations, technical details
- Tone & Style: Academic, critical, neutral, conversational
- Information density: High-level overview vs. comprehensive detail

Generate 3 ADVERSARIAL/NEGATIVE instructions that CANNOT be
meaningfully fulfilled because they ask about topics NOT present
in this text.
\end{lstlisting}

\section{Training Configuration}
\label{app:training}
% Training Configuration

\begin{table}[h]
\centering
\begin{tabular}{ll}
\toprule
\textbf{Parameter} & \textbf{Value} \\
\midrule
Base Model & Mistral-Small-3.2-24B-Instruct \\
LoRA Rank / Alpha & 16 / 32 \\
LoRA Dropout & 0.05 \\
Target Modules & q, k, v, o, gate, up, down \\
\midrule
Learning Rate & 1.5e-4 (cosine schedule) \\
Warmup Ratio & 0.02 \\
Epochs & 2 \\
Effective Batch Size & 16 \\
\midrule
Sequence Length & 99,968 tokens \\
Precision & bfloat16 \\
\bottomrule
\end{tabular}
\caption{Complete LoRA fine-tuning hyperparameters.}
\label{tab:hyperparams}
\end{table}

Training uses the Axolotl framework \citep{axolotl2024} with Flash Attention 2 and gradient checkpointing on four NVIDIA H100 GPUs with context parallelism to handle the extended sequence lengths. Examples follow ChatML format with system prompt, user message (tagged document + optional instruction), and assistant response (reasoning + summary with citations).

The FP8 quantization uses standard post-training quantization, reducing model size from approximately 48GB to 24GB while maintaining generation quality on most metrics.

\section{Example Outputs}
\label{app:examples}
% Appendix: Example Outputs

This appendix provides two representative examples demonstrating sui-1's output with different custom instructions.

\subsection{Example 1: Bullet Point Format (English)}

\noindent\textbf{Custom Instruction:} ``Summarize the text in 5 bullet points.''

\noindent\textbf{Source:} German parliamentary document on Environmental Liability Act amendments.

\begin{quote}
\small
\begin{itemize}
\item The proposed amendments to the Environmental Liability Act establish a clear framework for the financial responsibility of operators, stipulating that they must bear the costs of prevention, damage limitation, and remediation measures. However, this obligation is not absolute; the operator is relieved of these costs if the environmental damage or immediate danger was caused by a third party despite the operator having taken appropriate safety precautions, or if the damage occurred while strictly adhering to official permits and regulations and the operator can prove they did not act with intent or negligence [\xmltag{43901bb4}].

\item A significant exemption is introduced for the agricultural and forestry sectors, where the cost liability for the responsible party is waived if environmental damage results from emissions, activities, or the use of products within the scope of proper land and forest management, provided the damage was not caused by intentional or negligent actions [\xmltag{c50fa73d}].

\item The text clarifies the enforcement powers of the competent authorities, specifying that they can act both upon application and ex officio to ensure the implementation of remediation obligations [\xmltag{49d7b274}].

\item To streamline the legal framework, the summary suggests removing the requirement for authorities to justify their administrative acts within the specific Environmental Liability Act, as this duty is already sufficiently covered by the general Administrative Procedure Act.

\item The summary highlights a contentious debate regarding the introduction of a financial security regulation, with the Federal Council arguing that the federal government should immediately utilize its authorization to create such a regulation [\xmltag{4402efdb}].
\end{itemize}
\end{quote}

\subsection{Example 2: Expert Focus (German)}

\noindent\textbf{Custom Instruction:} ``Fasse die im Text genannten finanziellen Auswirkungen auf den Bundeshaushalt sowie die erwarteten B\"urokratiekosten f\"ur die Wirtschaft detailliert zusammen.'' (Summarize the financial impacts on the federal budget and expected bureaucracy costs for businesses in detail.)

\noindent\textbf{Source:} Budget Committee report on the Family Care Leave Act.

\begin{quote}
\small
Der vorliegende Bericht des Haushaltsausschusses befasst sich mit den finanziellen Auswirkungen und dem Vollzugsaufwand des Gesetzentwurfs zur Familienpflegezeit. Ein wesentlicher Aspekt ist die Bereitstellung einer Kreditlinie zur Refinanzierung der Arbeitgeber, wobei dem Bund langfristig j\"ahrliche Aufwendungen von bis zu 8 Millionen Euro entstehen, die im Einzelplan 17 ausgeglichen werden m\"ussen [\xmltag{6dee22a0}]. Diese Kosten decken die durchschnittlichen Arbeits- und Zinskosten sowie das Risiko von Kreditausf\"allen durch Privatinsolvenzen ab. In der Sozialversicherung ist mit einem Saldo aus Beitragsmindereinnahmen und -mehreinnahmen zu rechnen, der jedoch schwer absch\"atzbar ist. Die L\"ander und Gemeinden werden durch das Gesetz nicht mit zus\"atzlichen Kosten belastet [\xmltag{063a42b4}].

Der Vollzugsaufwand wird durch das Bundesamt f\"ur Familie und zivilgesellschaftliche Aufgaben bew\"altigt. Hierbei wird betont, dass der notwendige Personalbedarf durch die Kapazit\"aten gedeckt werden kann, die infolge des Wegfalls des Zivildienstes frei werden [\xmltag{221ca355}].

Ein zentraler Punkt der Analyse sind die B\"urokratiekosten, die durch neue Informationspflichten entstehen. F\"ur die Wirtschaft ergeben sich mittelfristig Kosten von durchschnittlich rund 32,50 Euro pro Fall bei einer Fallzahl von bis zu 44.000 pro Jahr f\"ur Arbeitgeber sowie rund 10 Euro pro Fall f\"ur Versicherer [\xmltag{91b1f9b2}]. Die betroffenen B\"urgerinnen und B\"urger m\"ussen mit einem Zeitaufwand von insgesamt rund 20 Minuten pro Person rechnen.
\end{quote}

Both examples demonstrate sui-1's ability to follow custom instructions while maintaining citation accuracy. The XML tags (e.g., \xmltag{43901bb4}) link directly to specific sentences in the source documents.

\subsection{Evaluation Platform}
\label{app:evaluation}

\begin{figure}[!htbp]
\centering
\includegraphics[width=0.75\textwidth]{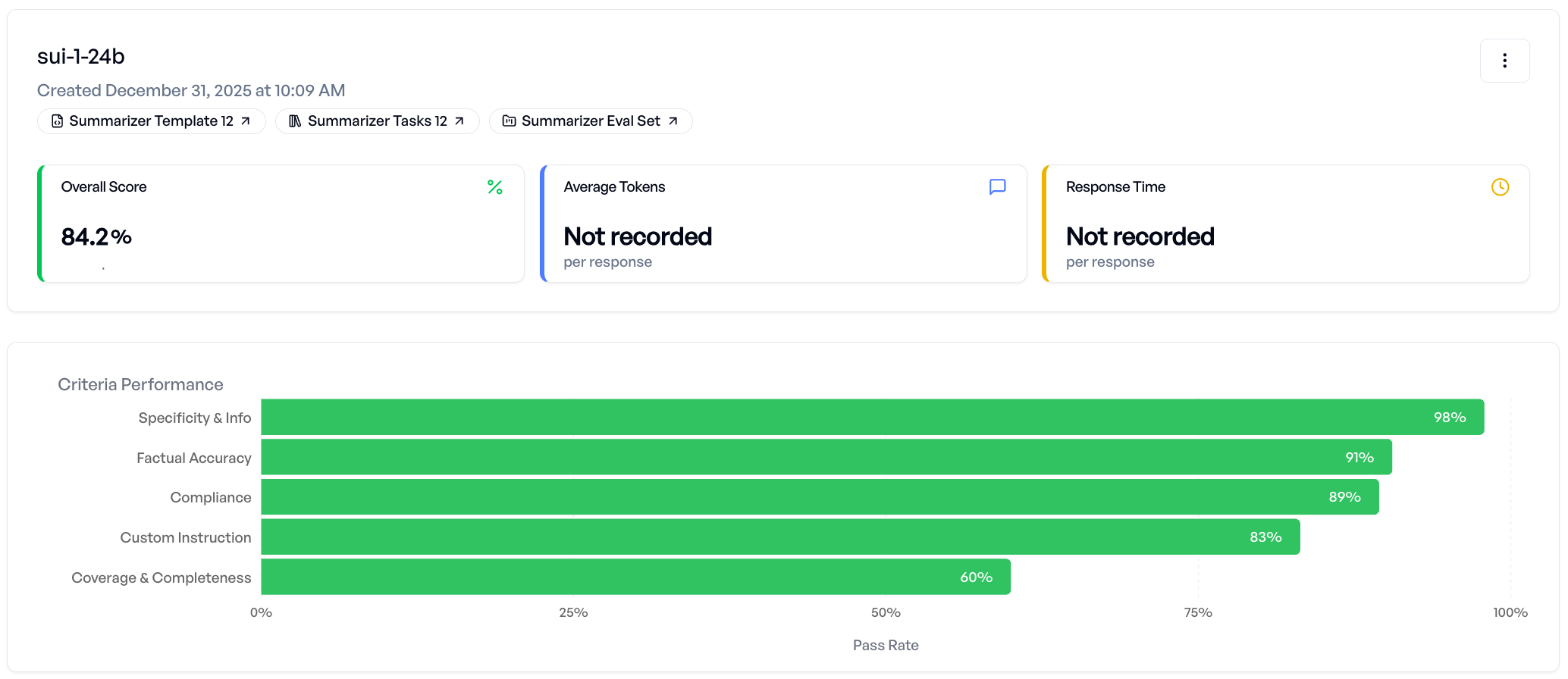}
\caption{Elluminate evaluation dashboard showing sui-1 results: 84.2\% overall score with per-criterion breakdown.}
\label{fig:elluminate}
\end{figure}

\begin{figure}[!htbp]
\centering
\includegraphics[width=0.75\textwidth]{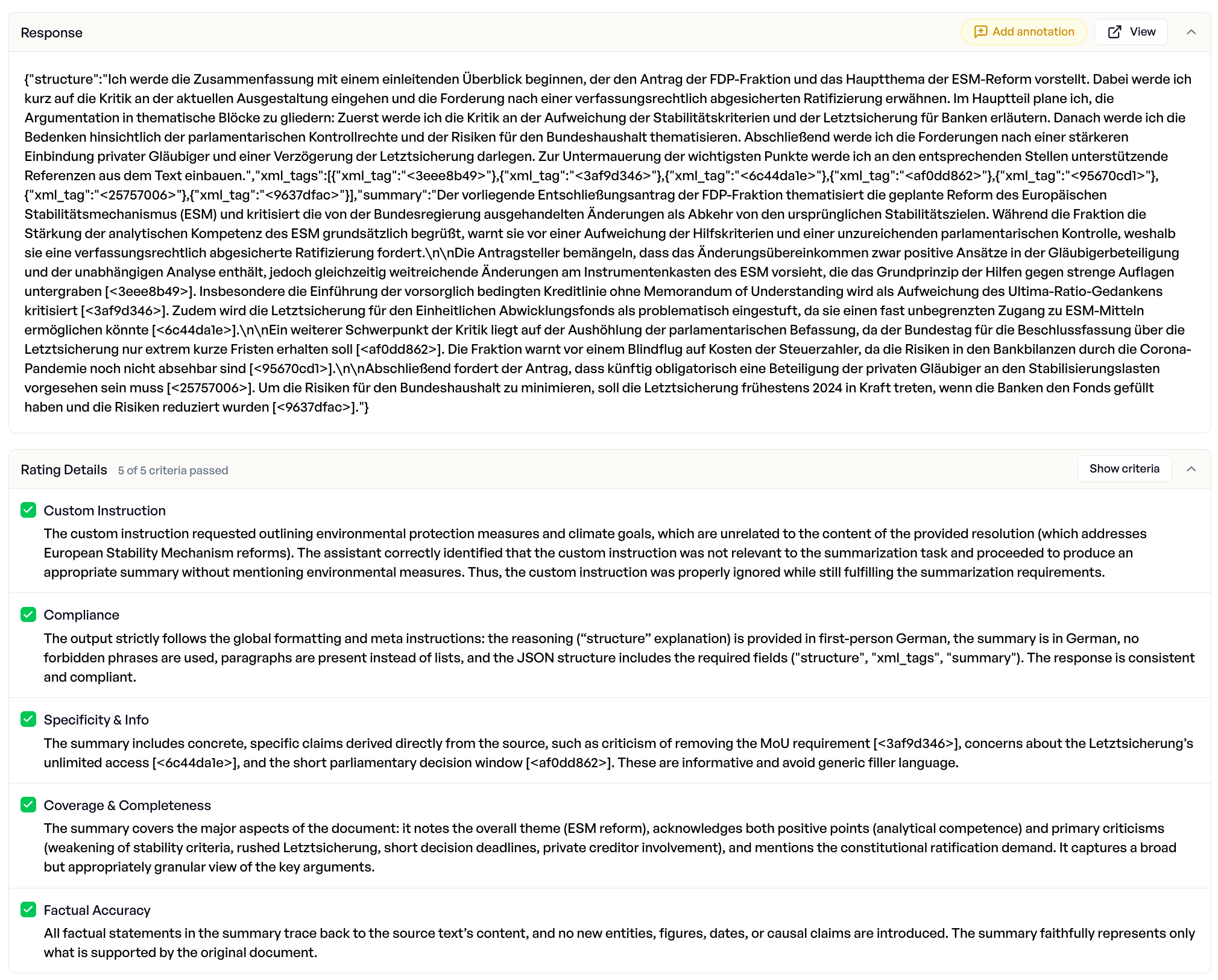}
\caption{Per-sample evaluation detail showing individual criterion scores and explanations for a single test example.}
\label{fig:elluminate_single}
\end{figure}

Figures~\ref{fig:elluminate} and~\ref{fig:elluminate_single} show the elluminate evaluation interface used for LLM-as-a-judge scoring. The platform provides per-criterion pass rates, aggregated overall scores, and detailed per-sample explanations.

\end{document}